\documentclass{article}
\UseRawInputEncoding
\usepackage[preprint]{neurips_2021}
\usepackage[utf8]{inputenc}
\usepackage{amssymb}
\usepackage{amsmath}
\usepackage{mathtools}
\usepackage{amsthm}
\usepackage[section]{placeins}
\usepackage{tikz}
\usepackage{dirtytalk}
\usepackage{wrapfig}
\usepackage{natbib}
\usepackage[linesnumbered,ruled,vlined]{algorithm2e}
\usepackage{neurips_2021}
\usepackage{graphicx}
\usepackage{tikz}
\usepackage[mode=buildnew]{standalone}
\usepackage{pythonhighlight}
\usetikzlibrary{shapes.geometric,positioning}
\usepackage{xcolor}
\usetikzlibrary{automata,positioning}
\title{
Log2NS: Enhancing Deep Learning Based Analysis of Logs With Formal to Prevent Survivorship Bias
}

\author{
\and Charanraj Thimmisetty
\and Praveen Tiwari
\and Didac Gil de la Iglesia
\and Nandini Ramanan
\and Marjorie Sayer
\and  Viswesh Ananthakrishnan
\and  Claudionor Nunes Coelho Jr\thanks{Corresponding author: \texttt{ccoelho@paloaltonetworks.com}}\\
\\
Advanced Applied AI Research \\
Palo Alto Networks \\
https://www.paloaltonetworks.com/}

\date{April 2021}

\begin{document}

\maketitle
\begin{abstract}
Analysis of large observational data sets generated by a reactive system is a common challenge in debugging system failures and determining their root cause. One of the major problems is that these observational data suffers from survivorship bias.  Examples include analyzing traffic logs from networks, and simulation logs from circuit design. In such applications, users want to detect non spurious correlations from observational data, and obtain actionable insights about them. In this paper, we introduce log to Neuro-symbolic (Log2NS), a framework that combines probabilistic analysis from machine learning (ML) techniques on observational data with certainties derived from symbolic reasoning on an underlying formal model. We apply the proposed framework to network traffic debugging by employing the following steps. To detect patterns in network logs, we first generate global embedding vector representations of entities such as IP addresses, ports, and applications. Next, we represent large log flow entries as clusters that make it easier for the user to visualize and detect interesting scenarios that will be further analyzed. To generalize these patterns, Log2NS provides an ability to query from static logs and correlation engines for positive instances, as well as formal reasoning for negative and unseen instances. By combining the strengths of deep learning and symbolic methods, Log2NS provides a very powerful reasoning and debugging tool for log-based data. Empirical evaluations on a real internal data set demonstrate the capabilities of Log2NS.

\end{abstract}

\section{Introduction}

\begin{figure} [htp!]
\begin{center}
\includegraphics[width=0.5\textwidth]{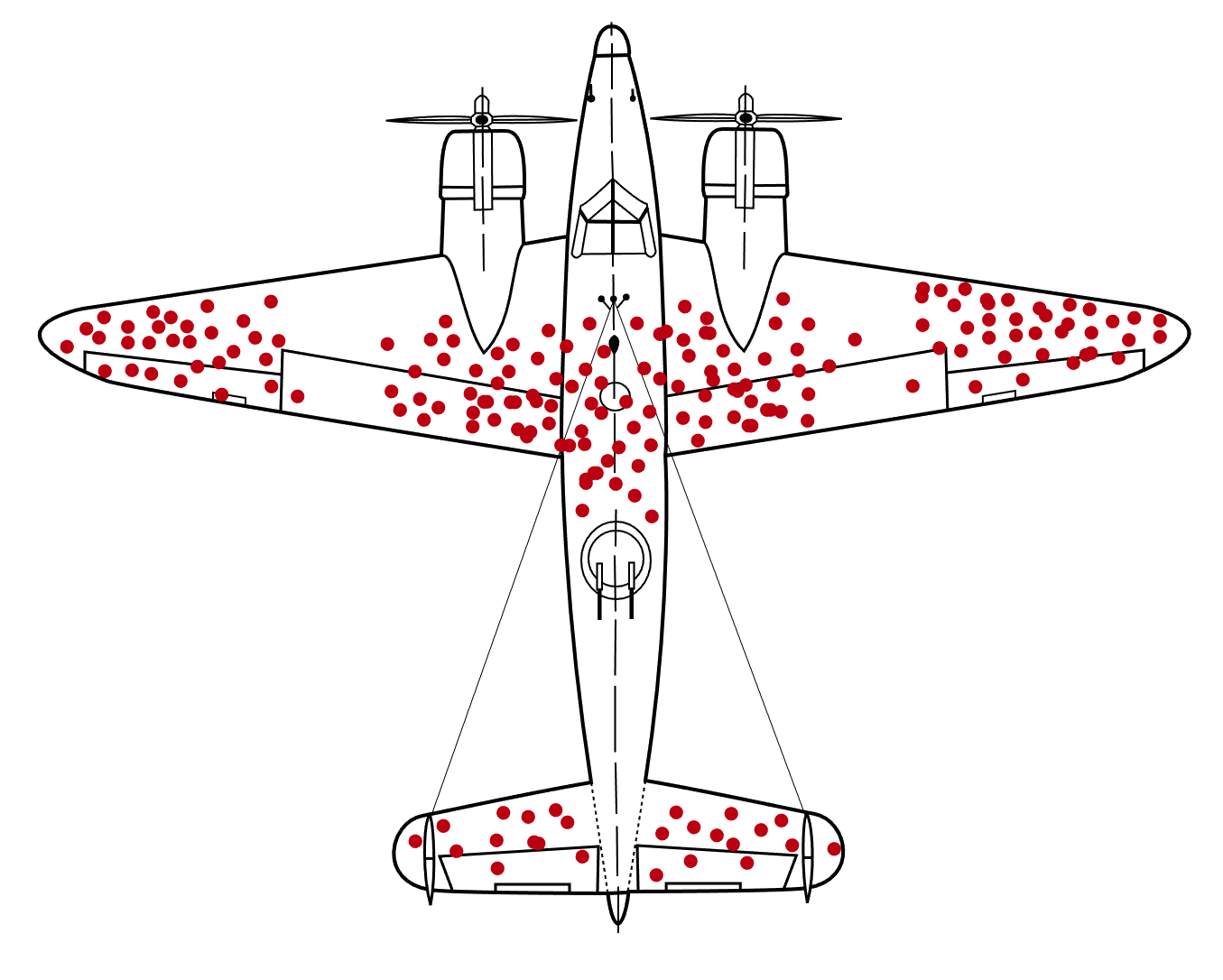}
\caption{Suvivorship bias}
\label{fig:survivorship-bias-airplane}
\end{center}
\end{figure}

Survivorship bias~\cite{wald1980reprint} refers to systematic error about our understanding of the world, where we analyze data only based on success cases, omitting consideration (on purpose or not) of the failing cases. One of the most known cases is pictured in Fig.~\ref{fig:survivorship-bias-airplane}~\footnote{credits to By Martin Grandjean (vector), McGeddon (picture), Cameron Moll (concept) - Own work, CC BY-SA 4.0, https://commons.wikimedia.org/w/index.php?curid=102017718} when Abraham Wald attempted to reduce bomber losses due to enemy fire in World War II.  During the analysis, he suggested that the bullet marks (hypothetically represented as red dots) showed only cases where the airplanes could land safely, whereas portions without red dots were due to cases when the bombers crashed. Abraham Wald recommended reinforcing portions without red dots, contrary to the original belief that the areas with red dots would need to be reinforced as they showed areas where enemy fire hit the bombers.

There are three takeaways from this example that we will address in this paper.

\begin{itemize}
    \item It is very easy to analyze complex scenarios through visualizations of observational data based on positive (red dots) and negative (lack of red dots) scenarios;
    \item Observational data will consist mostly of positive examples;
    \item If we cannot get insights from failures or negative scenarios, our analysis may be compromised.
\end{itemize}


By observational data, we are referring in this paper to data collected over time from a reactive system~\cite{harel1985development} - a system that responds to changing actions and conditions, influenced by logical rules and structures. The collected data consists of sequences of inputs and outputs of the system, or sometimes just the outputs, which we denote by {\it logs} of the system. We assume an associated formal mathematical model, sometimes referred to simply as the {\it model}. We assume here the formal model encompasses all of the rules and constraints of the reactive system.

We consider that both the real reactive system and the model are capable of generating log data. However, the formal model by construction is a conceptual representation of the real world, which may contain further restrictions on how the reactive system behaves over time. For example, in a real world we may have limitations on what inputs a reactive system may see in the future based on previous outputs, such as in the case of intelligent agents~\cite{sutton2011reinforcement}.

In numerous Machine Learning and Deep Learning based studies of log data ~\cite{dangut2020integrated, fontaine2019log, fang2020application, calabrese2020sophia}, it is common for a user who wants to better understand the underlying behavior of the system, or automatically analyze input-output responses of the system, to obtain actionable insights through the collected logs. However, by construction, these logs give rise to survivorship bias~\cite{shermer2014survivor}, simply because they reflect observable behaviors of the reactive system over a limited time and limited scenarios, with only  normal cases or with at most a very limited number of rare cases~\cite{mehta2020systemverilog, takakis2019dynamic}. Some of the representative examples where validation and analysis based on observational logs can lead to catastrophic consequences are the Pentium bug~\cite{pratt1995anatomy}, and the crash of flight AF-447~\cite{bottyan2010accident}.



When analyzing logs, several techniques have been used in the past to reduce the effect of this bias of observable data, such as the use of constrained random simulation~\cite{huang2021faster} or generation of synthetic training data~\cite{nikolenko}. However, it eventually becomes too hard if not impossible to collect data on certain rare scenarios, and even attempting to creating synthetic data may become a complex task~\cite{haixiang}.

Logs can consists of vast amounts of data, and it is usually useful to determine if one has seen anomalies in the logs, usually occurring as a post processing step~\cite{omar} or during data collection~\cite{mehta2020systemverilog}. Although machine learning models are useful to summarize and find anomalies in the data in a probabilistic way, because data is biased during data generation, a user may be interested to determine if a scenario is ever possible to occur. Although this question may not be suitable for a machine learning or deep learning model, it can be answered by a symbolic proof system based on a formal model of the reactive system~\cite{clarke, kropf2013introduction, gupta1992formal, bernardi2021security}. 

\begin{figure}[htb!] 
    \centering
    \begin{tikzpicture}[>=stealth]

  \node [draw, minimum width=12cm, minimum height=5cm] at (0,2) (A) {};
  \node [above=of A, yshift = 0.10cm, xshift = -3cm] (B) {\large{Logs}};
  \node [above=of A, yshift = -1.25 cm, xshift = -3cm] (C) {};
\node[above =of A, align=center, yshift = 0.10cm, xshift = 3cm](D) {\large{Configurations} \\ \large{(Models)}};
 \node [above=of A, yshift = -1.25 cm, xshift = 3cm] (E) {};
 \draw [->] (B) -- (C);
 \draw [->] (D) -- (E);
 \node[draw=black, align=center, fill=none, circle, minimum size=4cm, text width=3.2cm,very thick] (F) at (-3,2) {\textcolor{black}{{\Large\textbf{Statistical}}\\Machine learning\\Deep learning \\ Data mining \\ Probabilistic models}};
 \node[draw=black, align=center, circle,fill=none,inner sep = 0pt,minimum size=4cm, text width=3.2cm, very thick] (G) at (3,2) {\textcolor{black}{{\Large \textbf{Symbolic}}\\Formal logic\\Automated reasoning \\ Ontologies\\ Model checking}};
 \path
(G) edge [bend right, ultra thick, ->] node {} (F)
(F) edge [bend right, ultra thick, ->] node {} (G);
\node[yshift = -0.2cm](0,0)(H) {\textbf{Neural-Symbolic System}};
\node[below = of H](0,0)(I) {\large \textbf{Knowledge}};
\draw [->] (A) -- (I);
\end{tikzpicture}
    \caption{Revisit of figure from~\cite{roberta}}
    \label{fig:intro_1}
\end{figure}
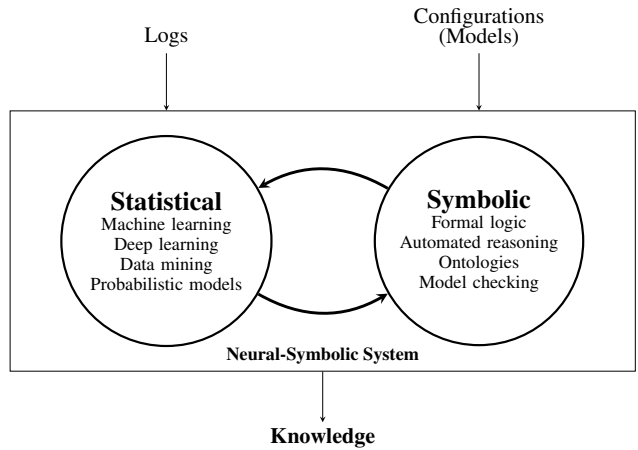

This paper presents {\it Log2NS (Log-to-neural-symbolic)}, a framework that enables users to reason about logs from a reactive systems by using machine learning and formal techniques. This solution can be best represented in Fig.~\ref{fig:intro_1} as a revisit of the figure originally presented at~\cite{roberta}, and it can be seen as an instance of using neural-symbolic systems~\cite{lamb, lamb2020graph}.
The framework first analyzes logs by computing embedding vectors on the entities of the logs. Then, it combines these vectors to represent log entry entities, using these representations to create visualizations. A powerful query language enables users to query the entire system for positive examples, by statically searching the logs or by computing correlations on the vector embeddings, and it uses formal engines to generate negative examples or even positive examples, when the query is only partially expressed or expressed as complex set of constraints.  In this case, the biased insights from observational data are complemented by queries using the formal engine, providing users with a comprehensive insights that enables one to draw conclusions.
The rest of the paper is organized as follows: Section 2 outlines the related work. In section 3 we will introduce the necessary background on Representation learning using embedding and its applications. Section 3 also introduces Formal methods on security policy. In section 4, we present our approach. Finally, in section 5, we conclude the paper by presenting our experimental evaluation on a real data and outlining the areas for future research.
\section{Related Work}
Machine learning has proved very powerful but has the limitation of interpretability. Symbolic AI was the main direction taken by AI research for a long time. Neuro-symbolic AI is a relatively new approach to AI that combines machine learning and symbolic AI. Neuro-symbolic AI can be seen as largely working within the framework of symbolic AI, using machine learning to infer and build the formal model upon which symbolic AI can reason \cite{lamb}. Our approach differs from Neuro-symbolic AI in that we begin with a formal model, and its observational data, and move forward from there. Rather than build a formal model from the ML model, our approach treats the ML model and formal model as collaborative players in a framework that accommodates complex real-world rule sets to enhance our understanding of the data flows they govern. 

The formal model studied in this paper begins with firewall configuration components and policy rules. These rules are built by people over time. They grow and accrete as their underlying networks concerns change. Maintenance of policy sets is challenging given network complexity and the changing landscape of threats. Of high importance are questions such as: do these rules keep my network secure? Is a particular rule letting in unwanted traffic? Are my rules too strict, and disallowing necessary traffic? Are the rules creating bottlenecks? Have route changes or network changes rendered my rules obsolete? Bodei et. al have developed methods to assess firewall configurations using a formal model~\cite{bodei2018language}. In Log2NS a machine learned model of the observational data and a formal model of the firewall configuration work together to increase understanding of possible network problems and potential flaws or inconsistencies in the firewall configuration. 

\section{Background}
\subsection{Representation Learning using Embedding}
The problem of analyzing logs for fault detection and diagnosis, particularly in security applications, is not new and has been a long rich area of research ~\cite{du2017deeplog,fu2009execution,Ring2019,lopez2017network,Ibrahim2016,Wei2009,Thomas2011}. To best automate log analysis using ML techniques, the first step is effective log representation. This section reviews a series of recently introduced methodologies from the field of representation learning that attempt to learn from complex high dimensional space by first transforming them into vectors, followed by a downstream learning task, e.g., say calculating behavioral similarities between network logs. 

Word2Vec is arguably the most popular word embedding model proposed by Mikolov et al~\cite{mikolov2013distributed} that project words to vectors of real numbers. Mikolov et al. proposed two different architectures; 1). Continuous bag of words (CBOW) neural network that maximizes the probability of a word given its context as in Eq. 1, and 2) Skip-gram model which uses a current word to predict the context words as in Eq. 2. 
\begin{eqnarray}
\frac{1}{\eta}\sum_{n=1}^{\eta}\log p(w_n|w_{n-\frac{c}{2}}...w_{n+\frac{c}{2}})\\
\frac{1}{\eta}\sum_{n=1}^{\eta}\sum_{m=n-c,m\neq n}^{n+c}\log p(w_m|w_n) 
\end{eqnarray} 
where $\eta$ is vocabulary size and $c$ is the size of context for each word. This also inspired some well known successful adaptation of Word2vec for nodes in large graph like Deepwalk ~\cite{perozzi2014deepwalk} and its extension node2vec~\cite{grover2016node2vec}. However, these methods suffer due to the inability to generalize to unseen data instances or to encode node attributes for graph embedding. Consequently, Hazem et al proposed a Graph Neural networks (GNNs) based method to perform graph representation learning at scale with better generalization~\cite{soliman2020graph}. The node embeddings are calculated using an information diffusion mechanism, where nodes broadcast information to their neighbours until convergence. With the advent of GNN, several graph-based representation techniques have emerged~\cite{hamilton2017inductive,peng2018large}. 
The most representative work in this line is Text-GCN by Yao et al. that achieves state-of-the-art results on benchmark domains~\cite{yao2019graph} with one big disadvantage of high memory utilization. 
Pennington et al. proposed GloVe, a method that combines the benefits of skip-gram and global matrix factorization. GloVe is based on the word-context co-occurrence matrix as:
\begin{align*}
    \mathcal{J} = f(\#(w_n,c_m))(\mathbf{w}_n^T \mathbf{c}_m + b_{w_{n}} + b_{c_{m}} - \log\#(w_n,c_m))^2
\end{align*} 
where $b_{w_{n}}$, $b_{c_{m}}$ are scalar biases for the target and context words. Ring et al. extended Word2vec and proposed IP2Vec that aims at transforming IP addresses into a continuous feature space $\mathbb{R}^m$ to compute distances between the IP addresses in this feature space~\cite{ring2017ip2vec}. They employ simple traffic descriptors as the context to train the Word2Vec model. Their empirical evaluations demonstrate the superiority of IP2Vec over other embedding techniques techniques within a botnet data set. In this paper, we will employ similar embeddings to learn meaningful vector representations of the network logs.

\subsection{Applications of Embedding}
Analysis and clustering of network traffic logs aim at answering several interesting questions in the domain of cybersecurity such as identifying host behaviors, grouping hosts with similar intentions, and so on. Unfortunately, this is an arduous task for humans to classify these logs that are generated in large volumes. Consequently, more often than not, this problem is posed as an unsupervised learning problem. Many approaches have been proposed to effectively condense or summarize these logs. These include, anomaly or outlier detection~\cite{Munz2007,Zhong2007,Wang2017,Wang2013}, network traffic classification~\cite{Pujari2017,Erman2006,Glennan2016,Singh2016}, and data security applications like policy generation, signature detection etc.~\cite{Finamore2011}, to name a few. 

Earlier systems heavily relied on rule-based approaches to classify or group log entries. These rules are limited to specific application scenarios and require extensive domain knowledge making the problem harder~\cite{cinque2012event,prewett2003analyzing,rouillard2004real}. Other generic methods typically apply a two-step procedure; first, parse log entries to structured forms (following the methodologies listed in section 3.1) and then employ the learning task say clustering. A clustering function $c$ takes a set of feature vectors as inputs and allocates them to appropriate clusters based on a similarity measure such that $c:\mathbb{R}^{|N|xd}\implies\mathbb{N}^{N}$. Choosing a clustering algorithm from the many options like Kmeans~\cite{macqueen1967some}, Gaussian Mixture Models~\cite{bishop2006mixture} etc. for the use case in hand is a non-trivial task. 

McGregor et al. proposed AutoClass, a probabilistic model-based clustering analysis to group logs using features from the transport layer~\cite{mcgregor2004flow}. Zander et al. proposed an enhancement of the Bayesian clustering technique using an expectation–maximization (EM) algorithm that helps determine the number of clusters as well as supports the soft clustering of the data.~\cite{zander2005automated}. However, the EM algorithm is often rather slow. In this work, we use the most efficient partition-based clustering method K-means which is a simple yet fast algorithm. For the set of feature vectors $\mathbb{X}\subseteq\mathbb{R}^d$, the k-means objective is to find a set $C={c_1,..c_k}$ of $k clusters$, such that it minimizes the average squared error $SSE = \sum_{x\in\mathbb{X}} min_{c\in C}(dist(x,c))$. Another successful method, DBSCAN, forms clusters based on the notion of density-reachability, i.e. A point is directly density-reachable to the objects in $\epsilon$-neighborhood of this point. However, the approach itself is very sensitive to the parameters. Erman et al. compared empirically the effectiveness of K-means,  DBSCAN, and AutoClass clustering for the traffic analysis and classification task, wherein the authors demonstrate the superiority of K-means over other state-of-the-art methods in learning high-purity clusters~\cite{Erman2006}.

\subsection{Formal methods on security policy}
Formal methods encompass a group of technologies for mathematical reasoning about the sanity of system behaviors. They have found successful applications in specifying, building and verifying complex software and hardware systems~\cite{d2008survey, stewart2021aadl, bernardi2021security, biere2021bounded}. A formal model of a given system is a precise mathematical description of its components and their relationships~\cite{edwards1997design}. Taken together, the model represents system behavior. Formal models are usually stated via mathematical formulae, often equations~\cite{barrett2010smt}. Formal engines~\cite{de2008z3} perform satisfiability checks to determine if the behavior of the system satisfies a given property, which also is described using a formal representation. A property in this context is a high level description of system behavior. If a solution exists, a trace depicting the solution is generated, otherwise engines return unsatisfiable i.e. no solution exists.

Formal synthesis/verification have proven to be effective in validation of firewall security policies ~\cite{liu2008formal, bodei2018language, beckett2017general}. A firewall configuration is composed of logical components, whose interactions define the firewall behavior. Security rules within the policy specify which traffic to accept or reject, based on filtering conditions. Policy administrators make use of a variety of filters such as IP addresses, geographical zones, applications etc. to specify security rules. Further, the specification enforces a precedence order amongst the rules. Examples of logical components in a configuration:
\begin{itemize}
\item network interfaces , zones, addresses, address groups
\item applications, application groups
\item address translation rules (NAT)
\item user, security profiles
\item routing tables 
\item security rules
\item address, address groups
\end{itemize}
To build a formal model from a firewall configuration, all its logical components and their relationships are converted to formulas.

\section{Our Approach}
\begin{figure}
    \centering
    \includegraphics[width=0.90\textwidth]{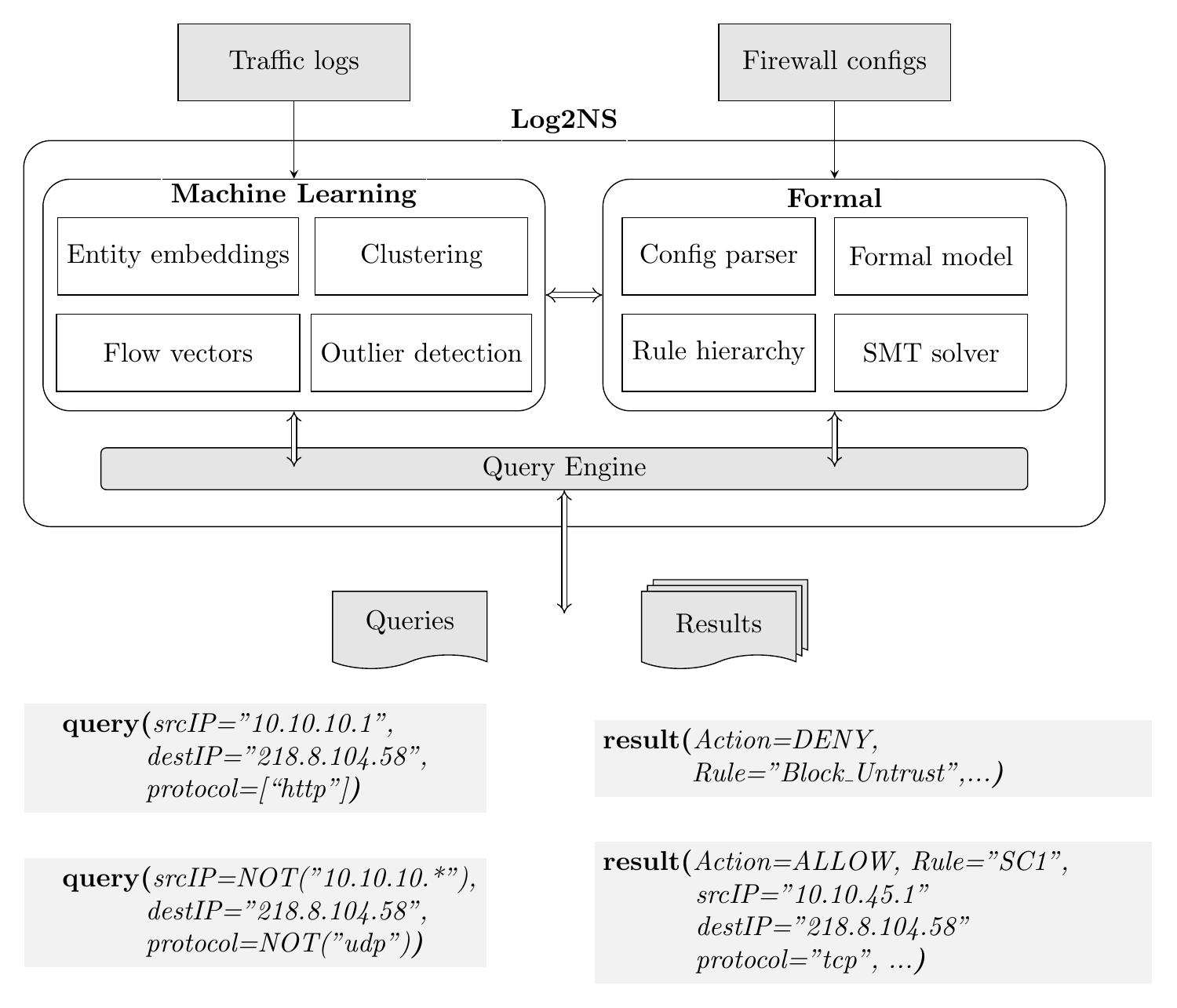}
    \caption{Log2NS architecture diagram}
    \label{fig:approach}
\end{figure}
Algorithm~\ref{algo:logsns} depicts the steps involved in Log2NS approach. It consists of two collaborative models, with a unified query interface for user interactions. As shown in Fig.~\ref{fig:approach}, the query engine routes user queries and interacts with the two models to return a final set of results. The Formal model is built by first extracting logical components from input firewall configuration files. Thereafter each logical component is converted to an equivalent formal representation. For example, Table~\ref{tab:1} shows a security rule (SR1) in firewall configuration and its conversion that allows traffic from \textit{Trust zone} to \textit{Internal zone} via any application. 

\begin{table}[h]
\centering
\caption{An example security rule}
\label{tab:1}
\begin{tabular}{lllll}
\hline\hline
Rule name & From Zone & Application & To Zone & Action\\
\hline
SR1       & Trust     & Any         & Internal & Allow \\
\hline
\end{tabular}
\end{table}
\begin{eqnarray}
\label{proposition1}
\text{From\_Zone} \in [\text{'Trust'}] \wedge \text{To\_Zone} \in [\text{'Internal'}] \rightarrow \text{Action} \in [1] \\
\label{zone_values}
\text{Zone} \in [\text{'Trust', 'Untrust', 'Internal'}] \\
\label{action_values}
\text{Action} \in [0,1] 
\end{eqnarray}


Entities like Source IP, Dest IP, Applications, Actions etc are considered as independent variables (Eq.~\ref{proposition1}). 
Some of these variables (as shown in Eq.~\ref{zone_values},~\ref{action_values}) can take values from a discrete set of possible values. The discrete set is inferred from the configuration. To provide solutions to user queries, an open source network configuration analysis tool is used~\cite{fogel2015general}. 

Firewall traffic logs are the other set of inputs to Log2NS for building the machine learning model. Traffic logs consist of a defined set of entities $e_1, e_2, \dots e_n$ ; example -  \textit{Source IP address, Destination IP address, IP protocol, Source port, Destination port} and \textit{Bytes sent}. Depending on the device and vendor these flow logs may consist of additional parameters such as \textit{From zone, To zone} and  \textit{Application}. Many of these entities are categorical and do not carry natural order; thus their meaning depends on context. To analyze logs we first create a global embedding of these entities via an approach similar to IP2Vec. Each log entry can be seen as a sentence as it carries meaning. For example, \textit{[10.0.0.1, 8.8.8.8, TCP, 80, 100]} states that a local machine is talking to Google server over port 80 using the TCP protocol and has sent 100 bytes of data. Log2NS defines context and target pairs from the set of log entities. Once the context-target pairs are generated for the input logs, the entity embedding gets computed by training a fully connected neural network with a single hidden layer whose size is much smaller than the input and output. The input of this network is a one-hot encoded context vector while the output is the probability of the target vector computed via a softmax function. The weights that map the input to the hidden layer give us the entity embedding. In practice, computing the probability of the target vectors becomes intractable due to the denominator of the softmax computation. There are two ways to handle this. One is to use negative sampling~\cite{mikolov2013distributed}, wherein the training step, instead of iterating over the entire vocabulary, we generate several negative examples from the noise distribution of the words. Because these negative examples have lower probability in the corpus, they simplify the softmax computation. Another approach is to use hierarchical softmax that uses an efficient binary tree format to represent the words in the vocabulary~\cite{mikolov2013efficient}. We use hierarchical softmax to compute the embedding for two reasons: 1. it works better for a corpus with infrequent words and in our case, some of the entity vectors might be present in only a few logs; 2. while generating the context-target pairs we ignore some of the possible combinations that appear in the same row (one observation logged from the firewall) as they should have zero impact on the embedding. Negative sampling might generate a few context-target pairs within the same row and assigning negative impact to these pairs might cause inaccurate embedding. 

Once the embeddings for the all the entities $e_1, e_2, \dots e_n$ are computed using the above procedure, we generate a vector for each row by concatenating the entity vectors. However, the concatenated vector might contain correlated entries. To remove the correlation and obtain a more compact low-dimensional representation, one can use non-linear manifold learning techniques such as kernel principal component analysis (KPCA)~\cite{scholkopf1997kernel} and diffusion maps~\cite{coifman2006diffusion}. Often times, using the correct distance metric to compute the correlation becomes important. Some of the manifold learning techniques such as diffusion maps allows us to measure intrinsic distance between the vectors~\cite{thimmisetty2014multiscale, thimmisetty2018high}. Instead of concatenation, one can also use weighted average to assign importance to some of the entities.  Once the vectors for each row is computed, we use unsupervised clustering techniques (K-means and Gaussian mixture models) to identify patterns in the logs.
     
\begin{algorithm}
\KwIn{logs, configuration\_file}
\KwOut{patterns\_in\_logs, positive\_examples, negative\_examples }
Generate context and target pairs \\
Compute the embedding of the log entities with hierarchical softmax \\
Generate vector for each row of the log entry \\
Create clusters using unsupervised learning \\ 
Create a formal model with configuration\_file\\
Identify the key question to be addressed \\
Use formal reasoning and log querying to generate positive and negative examples \\
\caption{Log2NS algorithm}
\label{algo:logsns}
\end{algorithm}
\newpage
\section{Experimental Evaluation}
\subsection{Data Description}
\begin{figure}[!htb] 
    \centering
    \includegraphics[width=.75\textwidth]{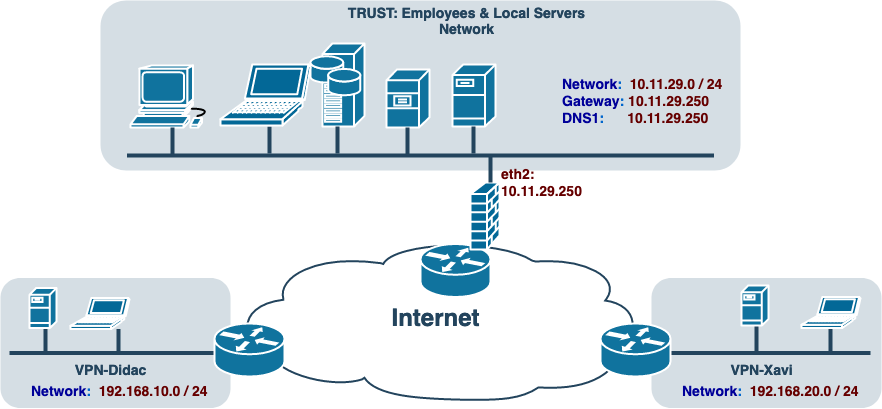}
    \caption{Network topology}
    \label{fig:topology}
\end{figure}
 In our experiments, we worked with network traffic data\footnote{Dataset will be provided as a supplemental material with Apache 2.0 licence} from a real network with the topology shown in Fig.~\ref{fig:topology}. The network has three sub networks of interest: a firewall-protected main office and two remote offices. The office firewall has IP address 10.11.29.250. The observational data spanned a period of three weeks. Raw flow records included source and destination IP addresses and ports, protocol, utilization, source and destination zones, identified applications, TCP flag and tunnel information. The network traffic is generated by users and by devices both inside and outside the network. Firewall configuration files during this period were also stored. 
\subsection{Experimental Setup} 
\label{sec:experiments}
We demonstrate the Log2NS framework by building ML model from the observational data, symbolic model from the firewall configuration. We show the effectiveness of our approach in analyzing the access patterns and associated security rules. 
\subsubsection{Building the Log Data Model}
The statistical model was created using three weeks of traffic data. Traffic log entries consisted of: \textit{Source Address, Destination Address, Application, IP Protocol, Source Region, Destination Region}. To generate embeddings, the context target pairs (Table~\ref{table:ctpairs}) were used. We used the embedding approach described earlier and implemented using the Gensim package~\cite{vrehuuvrek2011gensim}. We grouped the data into 20 clusters using K-Means~\cite{pedregosa2011scikit} and the model hyper parameters were obtained by grid search.
 
\begin{table}[h!]
\caption{Context-Target Pairs}
\centering
\begin{tabular}{c c}
\hline\hline
Context & Targets \\ 
\hline
Source Address & Destination Address, Application \\
Application & Destination Address \\
Destination Region & Source Address, Application \\
\hline
\end{tabular}
\label{table:ctpairs}
\end{table}

\subsubsection{Building the Formal Model}

To build the formal model, a previously generated  snapshot (same period as the traffic logs) of firewall configuration files were provided as inputs to Log2NS. 

\subsubsection{Generating Questions and Reasoning about our Models}
The next phase of the Log2NS framework is an iterative process: observations from the statistical model motivate queries to the symbolic model, which can lead to further observations and questions. From the clusters generated by the statistical model, we picked following positive examples of behavior to investigate: traffic to an unintended region, and dns queries outside of the firewall. 

a) Traffic to an unintended region was seen in the following cluster: 

\begin{table}[h!]
\caption{Unintended Region Cluster}
\centering
\begin{tabular}{c c c}
\hline\hline
Source IPs & Destination IPs & Applications \\ 
\hline
10.11.29.5 & 42.62.94.2 & \\
six additional & twenty-two additional&  not-applicable \\
source IPs & destination IPs & \\
 
\hline
\end{tabular}
\label{table:cluster_1}
\end{table}


This cluster stood out as its showed traffic to an unintended region. Further there was no application identified for the traffic. Through the unified query engine, Log2NS was able to confirm that this traffic is (and will be) allowed only by  BypassFw security rule (Table~\ref{table:smodel}). This exposed mis-configurations in policy, which could lead to security threats.


\begin{table}[h!]
\caption{Symbolic Model answers to Access to undesired region}
\centering
\begin{tabular}{l c c l}
\hline\hline
Filter Name &  Action & Line & Trace \\ [0.5ex]
\hline
&	  &  	&- Matched security rule BypassFW  \\
zone~Trust &	  &  	&- Matched source address \\
~vsys1~to& PERMIT& BypassFW& - Matched address any \\
zone~Untrust&  & & - Matched destination address \\
~vsys1 &  & & - Matched service application-default\\
& &  & - Matched application any\\
\hline
\end{tabular}
\label{table:smodel}
\end{table}

b) The next cluster of interest (Table~\ref{table:cluster_2}) showed source IP addresses that connected to internet dns servers (4.4.4.4 and 8.8.8.8).

\begin{table}[h!]
\caption{DNS Queries to Internet Cluster}
\centering
\begin{tabular}{c c c}
\hline\hline
Source IPs & Destination IPs & Applications \\ [0.5ex]
\hline
192.168.1.254 & 4.4.4.4 & \\[-0.5ex]
10.11.29.222 & 8.8.8.8 & dns \\
10.11.29.6 & & \\
\hline
\end{tabular}
\label{table:cluster_2}
\end{table}

We wanted to disable access to internet dns servers, such that dns requests use local dns server (10.11.29.250).
We added new security rules to disallow such accesses and validated the change with Log2NS. 

\subsection{Future directions}
\label{sec:future}
Building an accurate formal model from a specified firewall configuration is often the most critical and challenging step towards using such a model for property verification. The Log2NS framework will be enhanced to select a fixed number of traffic logs pertaining to security rules, to create formal ‘witness’ properties. Witness properties check for the existence of specified traffic scenarios against the formal model. Any failures will point to an over-constrained model, indicating nuances not captured during formal model creation. Further, Log2NS will be enhanced to perform incremental training of the machine learning model. The traces generated by the formal model would be used as pseudo input logs for training.  
\section{Conclusion}
\label{sec:conclusion}
In this paper, we addressed the problem of enhancing insights from observational data coming from logs by using formal engines.  We showed use cases, where logs can suffer from the survivorship bias problem, and because of that, insights derived from them will be inherently biased, if analysis is not complemented by other methods.

We introduced Log2NS, a framework that enables users to reason about logs and to understand a complex system by using deep learning techniques on logs. We captured system behavior using embeddings on log entries, that were later combined, and clustering techniques were used to understand interesting scenarios. Because log data is biased, we enhanced the logs using formal technology. Formal engines and a constraint language that accepted partially specified constraints and negation enabled a user to query the framework to understand rare conditions or negative examples. We showed how the framework was used to extract insights in a network security environment, where logs were obtained from firewall network and security configuration. A corresponding formal model was used to determine if traffic could be accepted or denied, and explore optimizations of security rules.
\bibliographystyle{authordate1}
\bibliography{references}
\end{document}